\pdfoutput=1

\documentclass[11pt]{article}

\usepackage[]{ACL2023}

\usepackage{times}
\usepackage{latexsym}

\usepackage[T1]{fontenc}

\usepackage[utf8]{inputenc}

\usepackage{microtype}

\usepackage{inconsolata}

\usepackage{amsmath}
\usepackage{amssymb}
\usepackage{nccmath}

\usepackage{xcolor,colortbl}
\usepackage{adjustbox}
\usepackage{caption}
\usepackage{subcaption}
\usepackage{booktabs}

\usepackage{CJKutf8}

\title{Improving Self-training for Cross-lingual Named Entity Recognition \\ with Contrastive and Prototype Learning}

\newcommand*{\affaddr}[1]{#1} 
\newcommand*{\affmark}[1][*]{\textsuperscript{#1}}
\newcommand*{\email}[1]{\text{#1}}

\author{Ran Zhou\thanks{~~Ran Zhou is under the Joint Ph.D. Program between Alibaba and Nanyang Technological University.}\affmark[~~1,2]~~~Xin Li\affmark[1]\thanks{~~Corresponding author}~~~~Lidong Bing\affmark[1]~~~Erik Cambria\affmark[2]~~~Chunyan Miao\affmark[2]\\
	\affaddr{\affmark[1]DAMO Academy, Alibaba Group}\quad
	\affaddr{\affmark[2]Nanyang Technological University, Singapore}\\
	\email{{\tt\{ran.zhou, xinting.lx, l.bing\}@alibaba-inc.com}}\\
	\email{{\tt\{cambria, ascymiao\}@ntu.edu.sg}} \\
	}
 
\begin{document}
\maketitle
\begin{abstract}

In cross-lingual named entity recognition (NER), self-training is commonly used to bridge the linguistic gap by training on pseudo-labeled target-language data. However, due to sub-optimal performance on target languages, the pseudo labels are often noisy and limit the overall performance. 
In this work, we aim to improve self-training for cross-lingual NER by combining representation learning and pseudo label refinement in one coherent framework.
Our proposed method, namely ContProto mainly comprises two components: (1) contrastive self-training and (2) prototype-based pseudo-labeling. 
Our contrastive self-training facilitates span classification by separating clusters of different classes, and enhances cross-lingual transferability by producing closely-aligned representations between the source and target language. 
Meanwhile, prototype-based pseudo-labeling effectively improves the accuracy of pseudo labels during training. 
We evaluate ContProto on multiple transfer pairs, and experimental results show our method brings in substantial improvements over current state-of-the-art methods. \footnote{Our code is available at \url{https://github.com/DAMO-NLP-SG/ContProto}.}
\end{abstract}

\section{Introduction}



Cross-lingual named entity recognition (NER) \citep{tsai-etal-2016-cross, xie-etal-2018-neural} has seen substantial performance improvement since the emergence of large-scale multilingual pretrained language models \citep{devlin-etal-2019-bert, conneau-etal-2020-unsupervised}.
However, there is still a noticeable gap between zero-shot cross-lingual transfer and monolingual NER models trained with target-language labeled data.
To further bridge the linguistic gap between the source and target language, self-training is widely adopted to exploit the abundant language-specific information in unlabeled target-language data \cite{ijcai2020-543, yehai-etal-2020-feature, chen-etal-2021-advpicker}. In general, self-training (sometimes referred to as teacher-student learning \citep{wu-etal-2020-single}) uses a weak tagger (i.e. teacher model) trained on source-language data to assign pseudo labels onto unlabeled target-language data, which is then combined with labeled source-language data to train the final model (i.e. student model). Nevertheless, due to sub-optimal performances on target languages, the pseudo-labeled data contains a large number of errors and might limit the performances of NER models trained on them.

To optimize self-training for cross-lingual NER, several methods have been proposed to improve the quality of pseudo labels. 
One line of work focuses on selecting curated pseudo-labeled data for self-training via reinforcement learning \citep{liang2021reinforced} or an adversarial discriminator \citep{chen-etal-2021-advpicker}.
However, they do not fully utilize all the unlabeled data available.
\citet{wu-etal-2020-single,ijcai2020-543} exploit the full unlabeled dataset and alleviate the noise in pseudo labels by aggregating predictions from multiple teacher models. 
Likewise, \citet{liang2021reinforced} develop multi-round self-training which iteratively re-trains the teacher model to generate more accurate pseudo-labels.
Despite their effectiveness, both multi-teacher and multi-round self-training impose a large computational overhead.
Furthermore, the aforementioned methods are mostly data-driven and ignore the explicit modeling of cross-lingual alignment in the representation space.

In this work, we take a different approach and propose ContProto as a novel self-training framework for cross-lingual NER.
Unlike existing data selection methods, ContProto sufficiently leverages knowledge from all available unlabeled target-language data.
Compared with multi-teacher or multi-round self-training, our method improves pseudo label quality without training separate models.
Moreover, we explicitly align the representations of source and target languages to enhance the model's cross-lingual transferability.
Specifically, ContProto comprises two key components, namely contrastive self-training and prototype-based pseudo-labeling. 
Firstly, we introduce a contrastive objective for cross-lingual NER self-training.
Whereas typical supervised contrastive learning \citep{NEURIPS2020_d89a66c7} treats labeled entities of the same class as positive pairs, we further construct pseudo-positive pairs comprising of a labeled source-language entity and a target-language span predicted as the same entity type by the current model. 
Hence, such contrastive objective not only separates different entity classes for easier classification, but also better aligns representations of the source and target language, achieving enhanced cross-lingual transferability. 
Secondly, we propose a prototype-based pseudo-labeling to refine pseudo labels on-the-fly at each training step. We start with constructing class-specific prototypes based on the representations produced by contrastive self-training, which can be regarded as cluster centroids of each entity type. Then, by ranking the distances between the representation of an unlabeled span and each prototype, we gradually shift its soft pseudo label towards the closest class. As a result, errors in pseudo labels are dynamically corrected during training.

It is noteworthy that our contrastive self-training and prototype-based pseudo-labeling are mutually beneficial. On one hand, entity clusters generated by contrastive learning make it easier to determine the closest prototype and update pseudo labels correctly. 
In turn, the model trained on the refined pseudo labels becomes more accurate when classifying unlabeled spans, and yields more reliable positive pairs for contrastive learning.


Our contributions are summarized as follows:
(1) The proposed ContProto shows competitive cross-lingual NER performance, establishing new state-of-the-art results on most of the evaluated cross-lingual transfer pairs (five out of six). 
(2) Our contrastive self-training produces well-separated clusters of representations for each class to facilitate classification, and also aligns the source and target language to achieve improved cross-lingual transferability.
(3) Our prototype-based pseudo-labeling effectively denoises pseudo-labeled data and greatly boosts the self-training performance.

\section{Preliminaries}

\subsection{Problem Definition}

Cross-lingual named entity recognition aims to train a NER model with labeled data in a source language, and evaluate it on test data in target languages.
Following previous works \citep{jiang-etal-2020-generalizing,ouchi-etal-2020-instance,yu-etal-2020-named,li-etal-2020-unified,fu-etal-2021-spanner}, we formulate named entity recognition as a span prediction task. 
Given a sentence $X=\{x_{1},x_{2},...,x_{n}\}$, we aim to extract every named entity $e_{jk}=\{x_{j},x_{j+1},...,x_{k}\}$ and correctly classify it as entity type $y$.
Under zero-shot settings, labeled data $D_{l}^{src}$ is only available in the source language ($src$), and we leverage unlabeled data $D_{ul}^{tgt}$ of the target language ($tgt$) during training.

\subsection{Span-based NER}
\label{ssec:spanner}

Following \citet{fu-etal-2021-spanner}, we use the span-based NER model below as our base model.
Firstly, the input sentence $X=\{x_{1},...,x_{n}\}$ is fed through a pretrained language model to obtain its last layer representations $h=\{h_{1},...,h_{n}\}$. 
Then, we enumerate all possible spans $s_{jk}=\{x_{j},...,x_{k}\}$ where $1 \leq j \leq k \leq n$, to obtain the total set of spans $S(X)$. 
The representation for each span $s_{jk} \in S(X)$ can be the concatenation of the last hidden states of its start and end tokens $[h_{j};h_{k}]$. 
We additionally introduce a span length embedding $l_{k-j}$, which is obtained by looking up the (k-j)\textsuperscript{th} row of a learnable span length embedding matrix. 
Thus, we obtain the final representation of $s_{jk}$ as $z_{jk}=[h_{j};h_{k};l_{k-j}]$.
Finally, the span representation is passed through a linear classifier to obtain its probability distribution $P_{\theta}(s_{jk}) \in \mathbb{R}^{|\mathbb{C}|}$, where $\mathbb{C}$ is the label set comprising of predefined entity types and an ``O'' class for non-entity spans.

\subsection{Self-training for NER}

Typically, self-training (or teacher-student learning) for cross-lingual NER first trains a teacher model $\mathcal{M}(\theta_{t})$
on the available source-language labeled dataset $D_{l}^{src}$ using a cross-entropy loss:
\begin{equation}
\label{eqn:L_src}
\small
L_{src}=-\frac{1}{N} \sum_{X \in D_{l}^{src}} \frac{1}{|S(X)|} \sum_{s_{jk} \in S(X)} \sum_{c \in \mathbb{C}} y_{jk}^{c} ~\mathrm{log} P_{\theta_{t}}^{c}(s_{jk})
\end{equation}
where $N$ is the batch size, $y_{jk}^{c}=1$ for the true label of span $s_{jk}$ and $0$ otherwise.

Given an unlabeled target-language sentence $X \in D_{ul}^{tgt}$,
the teacher model then assigns soft pseudo label $\hat{y}_{jk} = P_{\theta_{t}}(s_{jk}) \in \mathbb{R}^{|\mathbb{C}|}$ to each span $s_{jk} \in X$.
The student model $\mathcal{M}(\theta_{s})$ will be trained on the pseudo-labeled target-language data as well, using a soft cross-entropy loss:
\begin{equation}
\label{eqn:L_tgt}
\small
L_{tgt}=-\frac{1}{N} \sum_{X \in D_{ul}^{tgt}} \frac{1}{|S(X)|} \sum_{s_{jk} \in S(X)} \sum_{c \in \mathbb{C}} \hat{y}_{jk}^{c} ~\mathrm{log} P_{\theta_{s}}^{c}(s_{jk})
\end{equation}

The total objective for the student model in vanilla self-training is:
\begin{equation}
L = L_{src} + L_{tgt}
\end{equation}

\begin{figure*}
    \centering
    \includegraphics[width=0.99\textwidth]{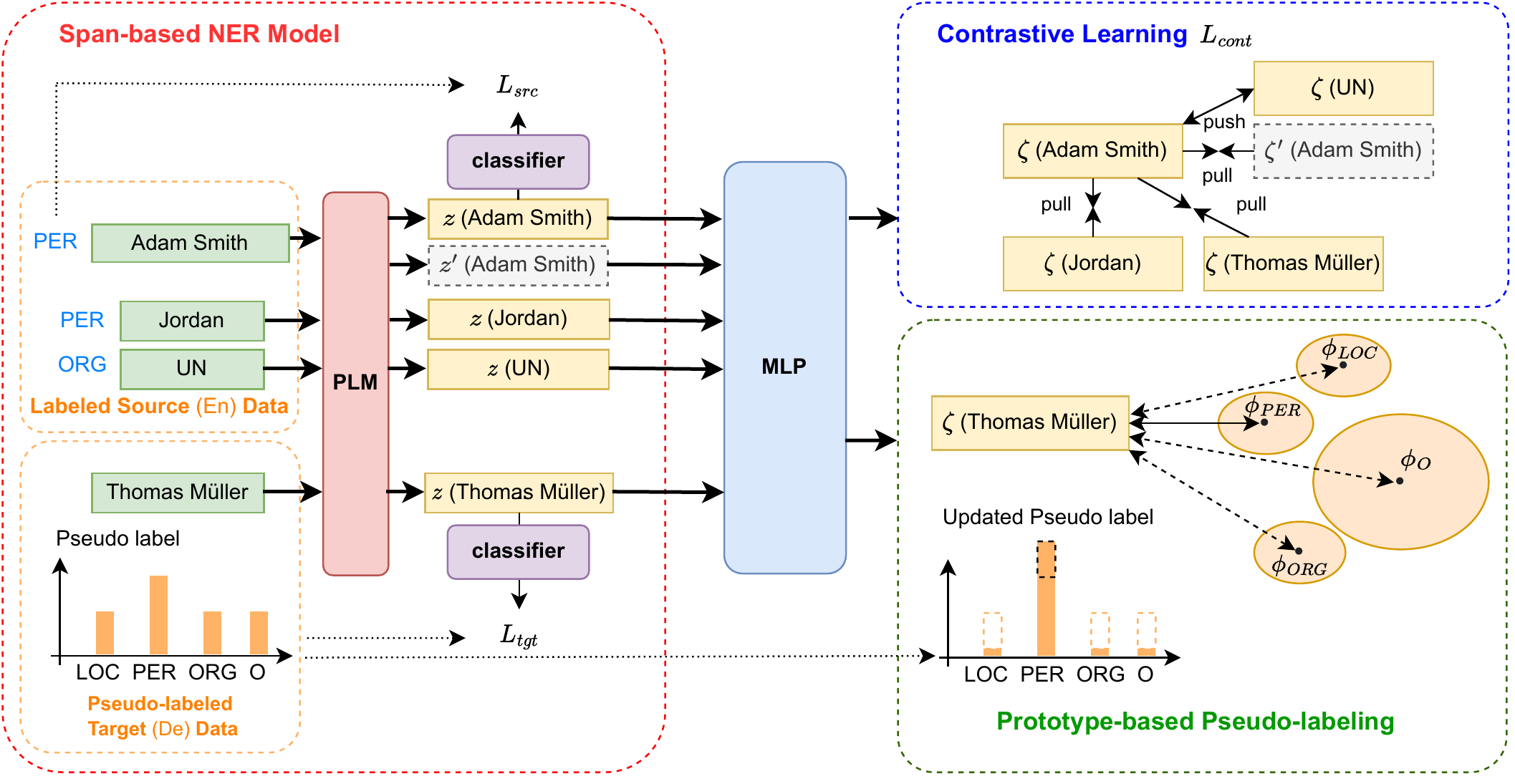}
    \caption{Illustration of ContProto. Both classifier blocks share the same parameters.}
    \label{fig:contproto}
\end{figure*}

\section{Methodology}

In this section, we present our self-training framework ContProto for cross-lingual NER.
As shown in the right part of Figure \ref{fig:contproto}, ContProto mainly comprises two parts, namely: (1) contrastive self-training (Section \ref{ssec:contrastive}) which improves span representations using contrastive learning; (2) prototype-based pseudo-labeling (Section \ref{ssec:prototype}) which gradually improves pseudo label quality with prototype learning. 

\subsection{Contrastive Self-training}
\label{ssec:contrastive}



In the following section, we first describe supervised contrastive learning for span-based NER, which focuses on source-language representations. Then, we introduce our pseudo-positive pairs, by which we aim to improve target-language representations as well. 

\paragraph{Supervised contrastive learning} 
We extend SupCon \cite{NEURIPS2020_d89a66c7} to span-based NER, which leverages label information to construct  positive pairs from samples of the same class and contrasts them against samples from other classes. 
Firstly, to generate multiple views of the same labeled source-language sentence, each batch is passed twice through the span-based NER model described in Section \ref{ssec:spanner}. 
An input sentence $X$ undergoes different random dropouts during each pass, such that each span $s_{jk} \in S(X)$ yields two representations $z_{jk}, z'_{jk}$. 
The span representations are further passed through a two-layer MLP, to obtain their projected representations $\zeta_{jk}, \zeta'_{jk}$. 
We denote the entire set of multi-viewed spans as $\{s_{i}, y_{i}, \zeta_{i}\}_{i=1}^{2m}$, where $y_{i}$ is the true label of $s_{i}$ and $m=\sum_{X} |S(X)|$ is the total number of spans in the original batch of sentences. 

Then, the supervised contrastive loss is defined as follows:
\begin{equation}
\label{eqn:L_cont}
\adjustbox{max width=0.49\textwidth}{
$L_{cont} = -\frac{1}{2m}\sum_{i=1}^{2m}\frac{1}{|P(i)|}\sum_{p \in P(i)} \mathrm{log}\frac{\mathrm{exp}(\zeta_{i} \cdot \zeta_{p} / \tau)}{\sum_{a \in A(i)} \mathrm{exp}(\zeta_{i} \cdot \zeta_{a} / \tau)}$
}
\end{equation}
where $A(i) \equiv \{1,2,...,2m\} \setminus \{i\}$, and $P(i) \equiv \{p \in A(i): y_{i}=y_{p}\}$ are indices of the positive sample set consisting of spans sharing the same label as $s_{i}$. Essentially, supervised contrastive learning helps to pull source-language entities of the same class together while pushing clusters of different classes apart, which induces a clustering effect and thereby benefits classification.

\paragraph{Pseudo-positive pairs}
As the aforementioned positive pair only involve source-language spans, it does not explicitly optimize target-language representations or promote cross-lingual alignment.
Therefore, we propose to construct pseudo-positive pairs which take target-language spans into account as well.

Concretely, we expand the multi-viewed
span set $\{s_{i}, y_{i}, \zeta_{i}\}_{i=1}^{2m}$ by adding in unlabeled target-language spans, where $m$ denotes the total number of spans from the source- and target-language sentences. 
For a source-language span, $y_{i}$ is still its gold label $y_{i}^{gold}$. However, as gold annotations are not available for target-language spans, we instead treat the model's prediction at the current training step as an approximation for its label $y_{i}$: 
\begin{equation}
\label{eqn:y}
y_{i}=
    \begin{cases}
    y_{i}^{gold} & \text{if $s_{i} \in D_{l}^{src}$}\\
    \mathrm{argmax}~P_{\theta}(s_{i}) & \text{if $s_{i} \in D_{ul}^{tgt}$}
    \end{cases}
\end{equation}
Likewise, we construct positive pairs from entities with the same (gold or approximated) label.
As an example, positive pairs for the \texttt{PER} (person) class might be composed of: (1) two source-language \texttt{PER} names; (2) one source-language \texttt{PER} name and one target-language span predicted as \texttt{PER}; (3) two target-language spans predicted as \texttt{PER}.
Therefore, apart from separating clusters of different classes, our contrastive self-training also explicitly enforces the alignment between languages, which facilitates cross-lingual transfer.

\paragraph{Consistency regularization}
We also include a consistency regularization term \citep{NEURIPS2021_5a66b920} to further enhance the model's robustness. Recall that each sentence is passed twice through the NER model, and each span $s_{i}$ yields two probability distributions $P_{\theta}(s_{i}), P'_{\theta}(s_{i})$ that are not exactly identical due to random dropout. Therefore, we enforce the model to output consistent predictions by minimizing the following KL divergence:
\begin{equation}
L_{reg} = -\frac{1}{m}\sum_{i=1}^{m}\textrm{KL}(P_{\theta}(s_{i})~||~P'_{\theta}(s_{i}))
\end{equation}

Finally, the total objective for ContProto is:
\begin{equation}
    L = L_{src} + L_{tgt} + L_{cont} + L_{reg}
\end{equation}

\subsection{Prototype-based Pseudo-labeling}
\label{ssec:prototype}

Benefiting from our contrastive self-training in Section \ref{ssec:contrastive}, entity representations (both source- and target-language) of the same class are tightly clustered together. 
Intuitively, the closest cluster to an unlabeled span is likely to represent the span's true class. 
Therefore, we can conveniently utilize these induced clusters as guidance to infer the unlabeled span's NER label.
To this end, we introduce prototype-based pseudo-labeling, which leverages prototype learning \citep{NIPS2017_cb8da676} to refine pseudo labels at each training step. 

\paragraph{Class-specific prototypes} To start off, we first define a series of prototypes $\phi_{c}$, each corresponding to a class $c \in \mathbb{C}$. A prototype $\phi_{c}$ is a representation vector that can be deemed as the cluster centroid of class $c$. 
Naively, $\phi_{c}$ can be calculated by averaging representations of class $c$ in the entire dataset at the end of an epoch. 
However, this means the prototypes will remain static during the next full epoch.
This is not ideal as distributions of span representations and clusters are vigorously changing, especially in the earlier epochs.
Hence, we adopt a moving-average style of calculating prototypes. 
Specifically, we iterate through a batch of mixed source- and target-language spans $\{s_{i}, y_{i}, \zeta_{i}\}_{i=1}^{m}$, 
and update prototype $\phi_{c}$ as the moving-average embedding for spans with (either gold or approximated) label $c$:
\begin{equation}
    \begin{split}
        \phi_{c} = \textrm{Normalize}~(\alpha \phi_{c} + (1-\alpha) \zeta_{i}), \\
        \forall i \in \{ i~|~y_{i} = c \}          
    \end{split}    
\end{equation}
Same as Equation \ref{eqn:y}, $y_{i}$ is either the gold label for source-language spans, or the approximated label obtained from the model's predictions for target-language spans. 
$\alpha$ is a hyperparameter controlling the update rate. 

\paragraph{Pseudo label refinement} Having obtained the prototypes, we then use them as references to refine the pseudo labels of target-language spans.
Typically, prototype learning classifies an unlabeled sample by finding the closest prototype, and assigning the corresponding label. 
However, this may cause two problems: (1) Assigning a hard one-hot label forfeits the advantages of using soft labels in self-training. (2) As the closest prototype might differ between consecutive epochs, there is too much perturbation in pseudo labels that makes training unstable. 
Thus, we again take a moving-average approach to incrementally update pseudo labels at each training step. Given a target-language span $\{s, \zeta\}$ at epoch $t$, its soft pseudo label from previous epoch $\hat{y}_{t-1}$ is updated as follows: 
\begin{equation}
\small
    \hat{y}^{c}_{t} = 
    \begin{cases}
        \beta \hat{y}^{c}_{t-1} + (1-\beta) & \text{if}~c = \arg \max_{\gamma \in \mathbb{C}} (\phi_{\gamma} \cdot \zeta) \\
        \beta \hat{y}^{c}_{t-1} & \text{otherwise}
    \end{cases}
\end{equation}
where $\hat{y}^{c}_{t}$ represents the pseudo probability on class $c$ and $\beta$ is a hyperparameter controlling the update rate. 
We use the dot product to calculate similarity $\phi_{\gamma} \cdot \zeta$, and define the distance between span representation and prototype as $(1-\phi_{\gamma} \cdot \zeta)$.
In other words, we find the prototype closest to the span's representation and take the corresponding class as an indication of the span's true label.
Then, we slightly shift the current pseudo label towards it, by placing extra probability mass on this class while deducting from other classes. 
Cumulatively, we are able to rectify pseudo labels whose most-probable class is incorrect, while reinforcing the confidence of correct pseudo labels. 

\paragraph{Margin-based criterion}

NER is a highly class-imbalanced task, where the majority of spans are non-entities (``O'').
As a result, non-entity span representations are widespread and as later shown in Section \ref{ssec:visualization}, the ``O'' cluster will be significantly larger than other entity types. 
Therefore, a non-entity span at the edge of the ``O'' cluster might actually be closer to an entity cluster. 
Consequently, the above prototype-based pseudo-labeling will wrongly shift its pseudo label towards the entity class and eventually result in a false positive instance.

To address this issue, we further add a margin-based criterion to enhance prototype learning. 
Intuitively, a true entity span should lie in the immediate vicinity of a certain prototype.
Thus, we do not update pseudo labels towards entity classes if the span is not close enough to any of the entity prototypes $\phi_{\gamma}$, i.e., the similarity between the prototype and any span representation $(\phi_{\gamma} \cdot \zeta_{i})$ does not exceed a margin $r$.
Meanwhile, as non-entity spans are widely distributed, we do not apply extra criteria and update a span as ``O'' as long as its closest prototype is $\phi_{O}$. 
Formally:
\begin{equation}
    \beta =
    \begin{cases}
        \beta & \text{if}~\arg \max_{\gamma \in \mathbb{C}} (\phi_{\gamma} \cdot \zeta_{i}) = \text{O} \\
        \beta & \text{if}~\max_{\gamma \in \mathbb{C} \setminus \{\text{O}\}} (\phi_{\gamma} \cdot \zeta_{i}) > r \\
        1 & \text{otherwise}
    \end{cases}
\end{equation}
We notice that different entity classes of different target languages might have varying cluster tightness, and thus it is not judicious to manually set a fixed margin $r$ universally. Instead, we automatically set class-specific margin $r_c$ from last epoch's statistics, by calculating the averaged similarity between target-language spans predicted as class $c$ and prototype $\phi_{c}$:
\begin{equation}
    r_{c} = \textsc{mean}(\phi_{c} \cdot \zeta_{i}), \text{where}~\arg \max P_{\theta}(s_{i}) = c
\end{equation}

Note that, at the start of training, our model does not produce well-separated clusters and the prototypes are randomly initialized. 
Therefore, we warm up the model by not updating pseudo labels in the first epoch. 

We highlight that our contrastive learning and prototype-based pseudo-labeling are mutually beneficial. 
By virtue of the clustering effect from contrastive learning, the resulting representations and prototypes act as guidance for refining pseudo labels. 
In turn, the model trained with refined pseudo-labels predicts unlabeled spans more accurately, and ensures the validity of pseudo-positive spans for contrastive learning.
To summarize, the two components work collaboratively to achieve the overall superior performance of ContProto.

\section{Experiments}

In this section, we verify the effectiveness of ContProto by conducting experiments on two public NER datasets with six cross-lingual transfer pairs and performing comparisons with various baseline models.

\subsection{Dataset}
Following previous works \citep{liang2021reinforced, li-etal-2022-unsupervised-multiple}, we evaluate our ContProto on six cross-lingual transfer pairs from two widely used NER datasets: 
(1) CoNLL dataset \citep{tjong-kim-sang-2002-introduction,tjong-kim-sang-de-meulder-2003-introduction}, which includes four languages, namely English (En), German (De), Spanish (Es) and Dutch (Nl); 
(2) WikiAnn dataset \citep{pan-etal-2017-cross} of English (En), Arabic (Ar), Hindi (Hi), and Chinese (Zh). 
Following common settings, we use the original English training set as our source-language training data $D_{l}^{src}$, while treating others as target languages and evaluate on their test sets. 
Annotations on target-language training sets are removed, and they are used as our unlabeled target-language data $D_{ul}^{tgt}$ for self-training. 
English development set is used for early stopping and model selection. 

\subsection{Baselines}

We mainly benchmark against the following self-training baselines for cross-lingual NER:

\noindent \textbf{TSL} \citep{wu-etal-2020-single} weights supervision from multiple teacher models based on a similarity measure as pseudo labels for self-training.

\noindent \textbf{Unitrans} \citep{ijcai2020-543} trains a series of teacher models sequentially using source-language data or translated data, and uses a voting scheme to aggregate pseudo labels from them.

\noindent \textbf{RIKD} \citep{liang2021reinforced} proposes a reinforced instance selector for picking unlabeled data and iteratively conducts self-training for multiple rounds.

\noindent \textbf{AdvPicker} \citep{chen-etal-2021-advpicker} leverages adversarial language discriminator for picking pseudo-labeled data.

\noindent \textbf{MTMT} \citep{li-etal-2022-unsupervised-multiple} introduces an extra entity similarity module and trains the student model with both NER and similarity pseudo labels.

We also compare ContProto with several baseline methods that do not leverage unlabeled target-language data, including Wiki \citep{tsai-etal-2016-cross}, WS \citep{ni-etal-2017-weakly}, TMP \citep{jain-etal-2019-entity}, BERT-f \citep{wu-dredze-2019-beto}, AdvCE \citep{keung-etal-2019-adversarial}, XLM-R\textsubscript{Large} \citep{pmlr-v119-hu20b}, mT5\textsubscript{XXL} \citep{xue-etal-2021-mt5}.

\subsection{Implementation Details}
We use XLM-R\textsubscript{Large} \citep{conneau-etal-2020-unsupervised} as the backbone pretrained language model of our span-based NER model. 
The dimension of the projected representations $\zeta_{i}$ for contrastive learning is set to 128. 
The model is trained for 10 epochs. AdamW \citep{loshchilov2018decoupled} is used for optimization and the learning rate is set to 1e-5.
We empirically set exponential moving average coefficients as $\alpha=0.99$ and $\beta=0.95$.
The batch size for both labeled source-language data and unlabeled target-language data is set to 16. 

\begin{table}[t!]
    \centering
    \resizebox{0.99\columnwidth}{!}{
    \begin{tabular}{lcccc}
    \toprule
         \textbf{Method} & \textbf{De} & \textbf{Es} & \textbf{Nl} & \textbf{Avg} \\
         \midrule
         \textit{w/o unlabeled data} \\
         \hspace{2mm} Wiki & 48.12 & 60.55 & 61.56& 56.74\\
         \hspace{2mm} WS & 58.50 & 65.10 & 65.40 & 63.00\\
         \hspace{2mm} TMP & 61.50 & 73.50 & 69.90 & 68.30\\
         \hspace{2mm} BERT-f & 69.56 & 74.96 & 77.57 & 74.03\\
         \hspace{2mm} AdvCE & 71.90 & 74.30 & 77.60 & 74.60\\
         \midrule
         \textit{self-training} \\
         \hspace{2mm} TSL & 75.33 & 78.00 & 81.33 & 78.22\\
         \hspace{2mm} Unitrans & 74.82 & 79.31 & 82.90 & 79.01\\
         \hspace{2mm} RIKD & \textbf{78.40} & 79.46 & 81.40 & 79.75\\
         \hspace{2mm} AdvPicker \\
         \hspace{5mm}\textit{- seq-tagging} & 75.01 & 79.00 & 82.90 & 78.97\\
         \hspace{5mm}\textit{- span-based}~\textsuperscript{\dag} & 73.93	& 84.70 &	81.01 &	79.88\\
         \hspace{2mm} MTMT & 76.80 & 81.82 & 83.41 & 80.68\\
         \midrule
         \hspace{1mm} \textbf{ContProto} \textit{(Ours)} & 76.41	& \textbf{85.02} & \textbf{83.69} & \textbf{81.71}\\
    \bottomrule      
    \end{tabular}
    }
    \caption{Experimental results on CoNLL. ContProto results are micro-F1 averaged over 3 runs.\textsuperscript{\dag}Implemented using span-based NER model. Baseline results without markers are cited from the original papers.}
    \label{tab:conll}
\end{table}

\begin{table}[t!]
    \centering
    \resizebox{0.99\columnwidth}{!}{
    \begin{tabular}{lcccc}
    \toprule
        \textbf{Method} & \textbf{Ar} & \textbf{Hi} & \textbf{Zh} & \textbf{Avg}\\
        \midrule
        \textit{w/o unlabeled data} \\
        \hspace{2mm} BERT-f  & 42.30 & 67.60 & 52.90 & 54.27\\
        \hspace{2mm} XLM-R\textsubscript{Large}   & 53.00 & 73.00 & 33.10 & 53.03\\
        \hspace{2mm} mT5\textsubscript{XXL}     & 66.20	& 77.80	& 56.80	& 66.93\\
        \midrule
        \textit{self-training} \\
        \hspace{2mm} TSL     & 50.91 & 72.48 & 31.14 & 51.51\\
        \hspace{2mm} RIKD    & 54.46 & 74.42 & 37.48 & 55.45\\
        \hspace{2mm} AdvPicker \\
        \hspace{5mm} \textit{- seq-tagging} \textsuperscript{\dag} & 53.76 & 73.69 & 41.24 & 56.23\\
        \hspace{5mm} \textit{- span-based} \textsuperscript{\ddag} & 70.70 & 80.37 & 56.57 & 69.21\\
        \hspace{2mm} MTMT    & 52.77 & 70.76 & 52.26 & 58.60\\    
        \midrule
        \hspace{2mm} \textbf{ContProto} \textit{(Ours)}  & \textbf{72.20}	& \textbf{83.45} & \textbf{61.47}	& \textbf{72.37}\\
    \bottomrule
    \end{tabular}
    }
    \caption{Experimental results on WikiAnn. ContProto results are micro-F1 averaged over 3 runs. \textsuperscript{\dag}Implemented using official source code. \textsuperscript{\ddag}Implemented using span-based NER model. Baseline results without markers are cited from the original papers.}
    \label{tab:wikiann}
\end{table}

\subsection{Main Results}

\paragraph{CoNLL results} We present the experimental results on CoNLL dataset in Table \ref{tab:conll}. Overall, our ContProto achieves the best results in terms of averaged F1 over the target languages, with a +1.03 improvement compared to the previous state-of-the-art MTMT. 
Compared with methods that do not use unlabeled data, ContProto presents substantial improvements, suggesting that incorporating target-language unlabeled data is indeed beneficial to cross-lingual NER.
Furthermore, our method outperforms both multi-teacher (i.e., TSL, Unitrans) and multi-round (i.e., Unitrans, RIKD) self-training. 
This shows our prototype learning produces more accurate pseudo labels compared to ensembling multiple teacher models or iterative self-training.
Compared with data selection methods (i.e., RIKD, AdvPicker), 
our superior performance demonstrates that on the premise of guaranteeing high-quality pseudo labels, it is beneficial to leverage as much target-language data as possible.

Although MTMT attempts to reduce the distance between entities of the same class in the same language, it does not account for the relation between a source- and a target-language entity.
Besides, AdvPicker implicitly aligns the source and target language during language-independent data selection but does not inherit those representations when training the final model. 
In comparison, our contrastive objective explicitly reduces the distance between a pair of source- and target-language entities of the same class, which aligns the source- and target-language representations to achieve better cross-lingual performance.

For a fair comparison, we further implement span-based NER based on the official codebase of AdvPicker~\citep{chen-etal-2021-advpicker}.
From experimental results, span-based AdvPicker shows some improvement over the original sequence tagging formulation. However, our ContProto still outperforms span-based AdvPicker by a considerable margin.

\paragraph{WikiAnn results}
As shown in Table \ref{tab:wikiann}, our ContProto achieves superior results on WikiAnn languages as well, with an averaged +3.16 F1 improvement compared to the best baseline method. 
It is noteworthy that span-based AdvPicker presents considerable improvements compared to its original sequence-tagging formulation, suggesting that span-based NER is a more appropriate formulation for identifying named entities in cross-language scenarios, especially for transfer pairs with larger linguistic gaps.
Compared with span-based AdvPicker, ContProto still shows a significant advantage by aligning source- and target-language representations and improving pseudo-label quality.

\begin{table*}[t!]
    \centering
    \resizebox{0.99\textwidth}{!}{
    \begin{tabular}{lllllll}
    \toprule
        \textbf{Method} & \textbf{De} & \textbf{Es} & \textbf{Nl} & \textbf{Ar} & \textbf{Hi} & \textbf{Zh}\\
        \midrule 
        \textbf{ContProto}  & \textbf{76.41}	& \textbf{85.02} & \textbf{83.69} & \textbf{72.20}	& \textbf{83.45} & \textbf{61.47}\\
        \hspace{3mm} \textit{- w/o proto} & 74.87 (-1.54)	& 84.08 (-0.94)	& 81.44 (-2.25)	& 71.49 (-0.71)	& 83.10 (-0.35)	& 59.57 (-1.90)\\
        \hspace{3mm} \textit{- w/o proto \& cl} & 74.17 (-2.24)	& 84.47	(-0.54)	& 81.03	(-2.66)	& 70.40	(-1.80)	& 81.00	(-2.45)	& 56.30	(-5.16)\\
        \hspace{3mm} \textit{- w/o reg} & 76.23 (-0.18)	& 84.96 (-0.06)	& 83.56 (-0.13)	& 72.15 (-0.05)	& 83.21 (-0.24)	& 61.31 (-0.16)\\
        \hspace{3mm} \textit{- fixed margin} & 74.65	(-1.76)	& 84.49	(-0.52)	& 83.09	(-0.60)	& 69.19	(-3.01)	& 83.07	(-0.38)	& 60.61	(-0.86)\\
        \hspace{3mm} \textit{- proto w/o cl} & 72.59	(-3.82)	&81.18	(-3.84)	&80.76	(-2.93)	&69.72	(-2.48)	&58.38	(-25.07)	&53.52	(-7.95)\\
    \bottomrule
    \end{tabular}
    }
    \caption{Ablation studies. Values in brackets indicate the performance drop compared to our full method.}
    \label{tab:ablation}
\end{table*}

\section{Analysis}

\subsection{Ablation Studies}

To demonstrate the contribution of each design component of ContProto, we conduct the following ablation studies: 
(1) \textit{w/o proto} which removes prototype-based pseudo-labeling and only keeps our contrastive self-training; 
(2) \textit{w/o proto \& cl} which removes both prototype-based pseudo-labeling and the contrastive objective; 
(3) \textit{w/o reg} which removes the consistency regularization; 
(4) \textit{fixed margin} which manually tunes a universally fixed margin $r=1.0$ instead of automatic class-specific margins; 
(5) \textit{proto w/o cl} which removes the contrastive objective, and directly uses the unprojected representation $z_{i}$ for constructing prototypes and updating pseudo labels.

Based on experimental results in Table \ref{tab:ablation}, we make the following observations:
(1) \textit{w/o proto} shows reduced performance on all target languages, which verifies the ability of our prototype-based pseudo-labeling in improving pseudo label quality.
(2) \textit{w/o proto \& cl} further lowers target-language performance, which demonstrates the effectiveness of contrastive self-training in separating different classes and aligning the source- and target-language representations. 
(3) \textit{w/o reg} demonstrates that removing the consistency regularization leads to slight performance drops on all target languages.
(4) Using a manually tuned universal margin, \textit{fixed margin} underperforms ContProto by a considerable amount. This signifies the flexibility brought by the automatic margin when cluster tightness differs between classes. 
(5) \textit{proto w/o cl} leads to drastic performance drops. Without the contrastive objective, clusters of different classes overlap with each other. As a result, the closest prototype might not accurately reflect a span's true label, and this leads to deteriorated pseudo label quality. Thus, the clustering effect from contrastive learning is essential for accurate prototype-based pseudo-labeling.

\begin{figure*}
     \centering
     \begin{subfigure}[b]{0.45\textwidth}
         \centering
         \includegraphics[width=\textwidth]{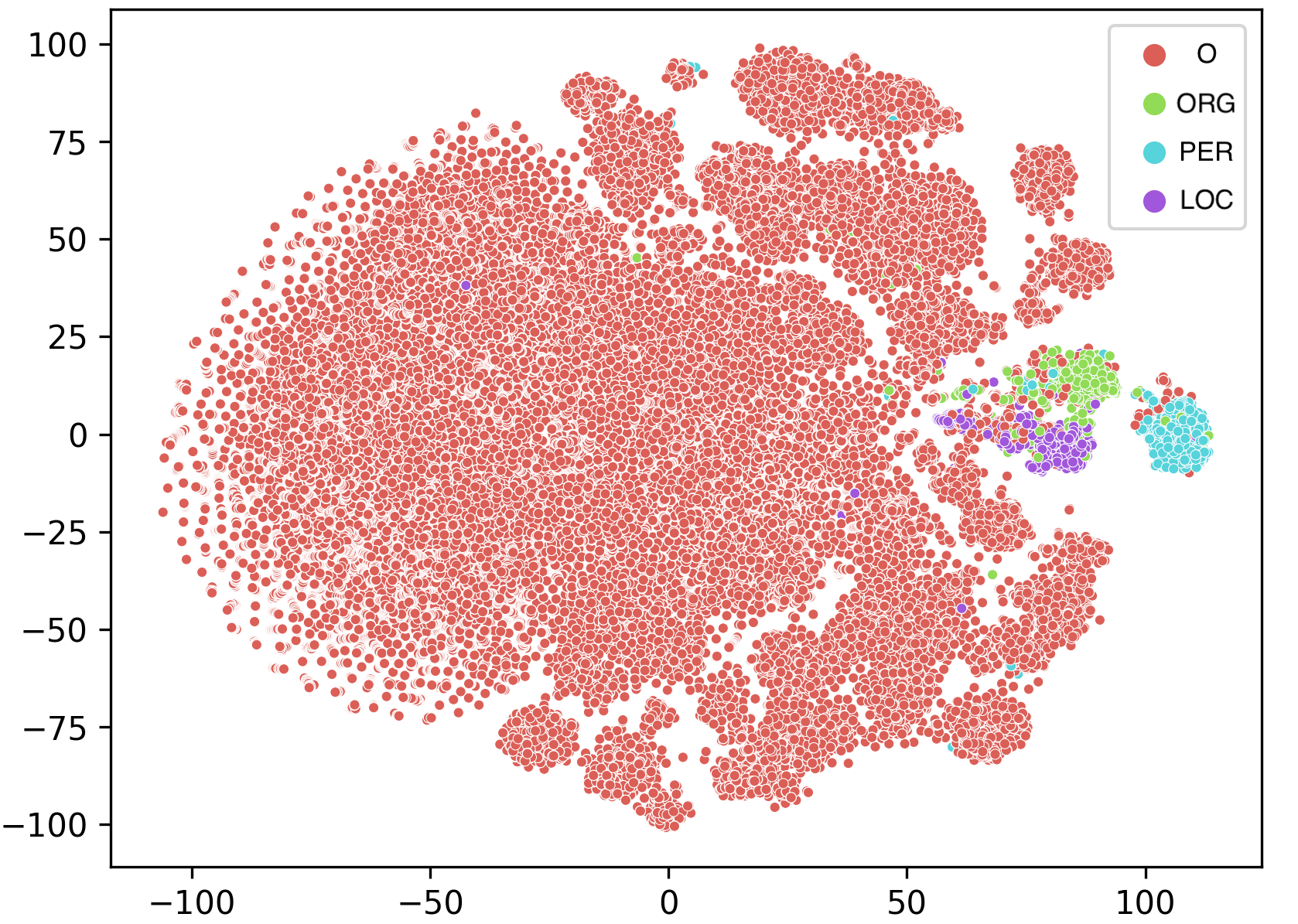}
         \caption{Vanilla self-training}
         \label{fig:tsne_a}
     \end{subfigure}
     \hfill
     \begin{subfigure}[b]{0.45\textwidth}
         \centering
         \includegraphics[width=\textwidth]{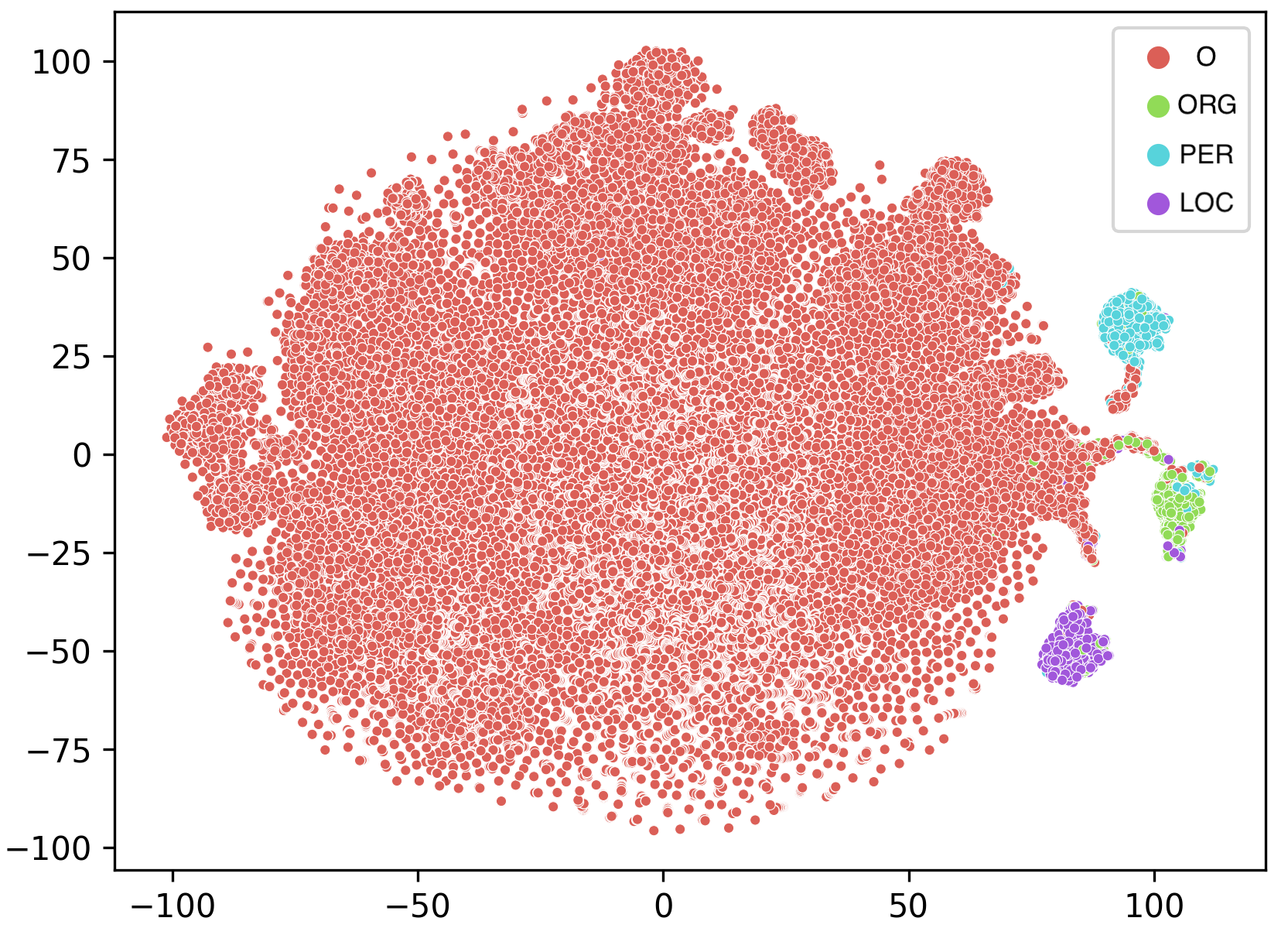}
         \caption{ContProto}
         \label{fig:tsne_b}
     \end{subfigure}
        \caption{t-SNE visualization of Chinese (Zh) spans.}
\end{figure*}

\begin{figure}[t!]
    \centering
    \includegraphics[width=0.49\textwidth]{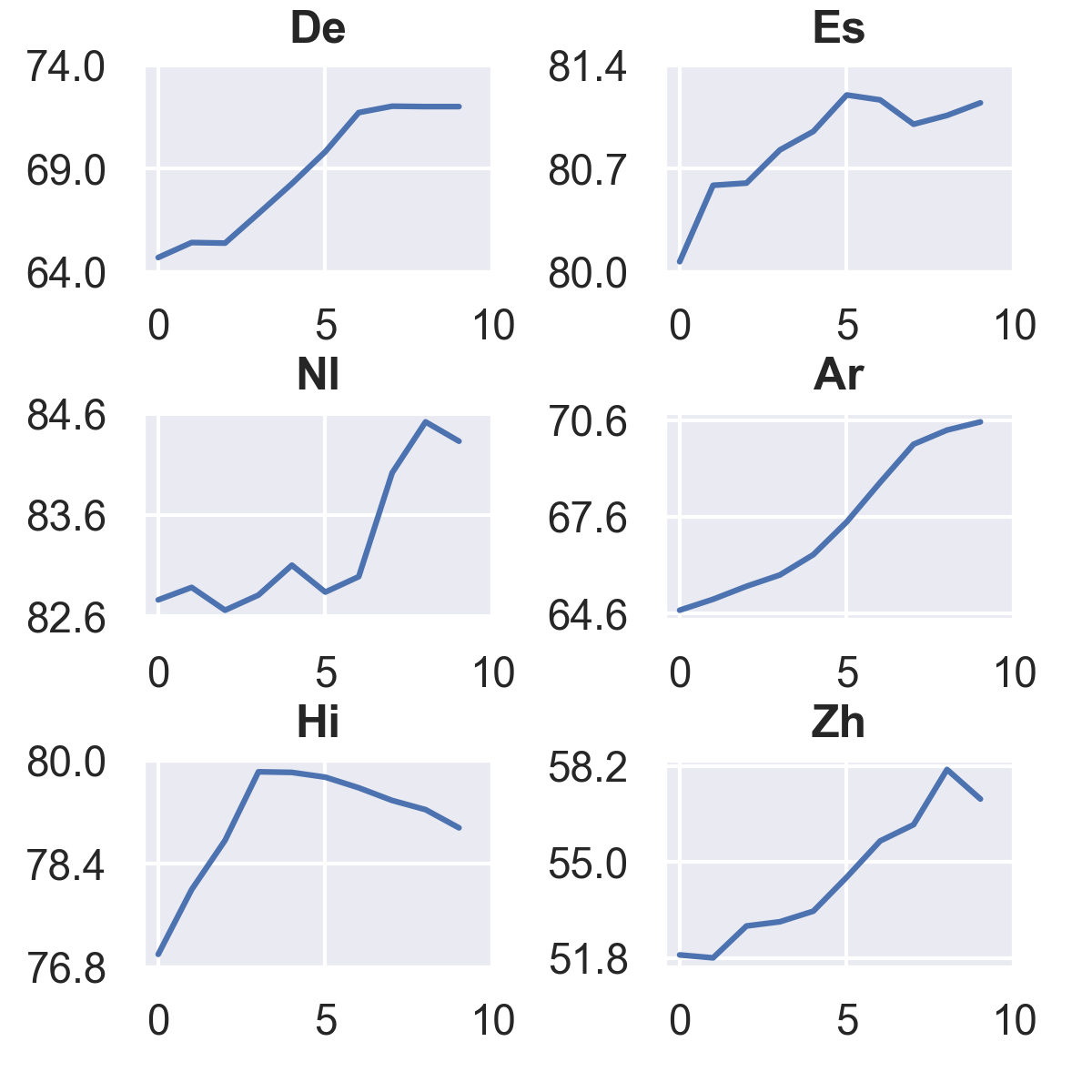}
    \caption{Pseudo label quality. The horizontal axis is the epoch number and the vertical axis is the oracle F1 of pseudo labels.}
    \label{fig:f1}
\end{figure}

\subsection{Visualizing Span Distributions}
\label{ssec:visualization}
We also conduct a t-SNE visualization \citep{van2008visualizing} of span representations $z_{i}$. 
As shown in Figure \ref{fig:tsne_a}, vanilla self-training generates representations with some overlapping between different classes, which makes it challenging to classify them. 
In contrast, our ContProto (Figure \ref{fig:tsne_b}) produces more distinguishable representations where clusters of different classes are separated, which verifies the effectiveness of our contrastive objective.
Furthermore, it can be easily seen that the non-entity ``O'' cluster is significantly larger than other entity classes, which justifies the necessity of margin-based criterion in Section \ref{ssec:prototype}.

\subsection{Pseudo Label Quality}

Recall that we remove gold labels from the original target-language training sets, and treat them as unlabeled data for self-training. For analysis purposes, we retrieve those gold labels, to investigate the efficacy of ContProto in improving the quality of pseudo labels.

Specifically, we take the gold labels as references to calculate the oracle F1 of pseudo labels at the end of each epoch.
As shown in Figure \ref{fig:f1}, the pseudo label F1 indeed improves during training on all target languages, proving the effectiveness of our prototype-based pseudo-labeling. 
Noticeably, there are significant F1 increases (5$\sim$7\%) on German (De), Arabic (ar), and Chinese (Zh).
On Hindi (Hi), however, we observe a slight drop of pseudo label F1 after epoch 3, which is mainly due to a reduction of pseudo label recall. We attribute this to the larger variance of Hindi entity distribution, such that many entities outside the automatic margin turn into false negatives. 
As the ablation study (\textit{w/o proto}) shows, prototype-based pseudo-labeling for Hindi only accounts for +0.35 performance improvement, and the overall improvement mainly comes from contrastive self-training. 
Still though, compared with initial pseudo labels, the updated Hindi pseudo label quality is improved.

\section{Related Work}
\paragraph{Cross-lingual NER} Existing methods for NER \citep{ding-etal-2020-daga, xu-etal-2021-better, xu2022clozing,xu2022peerda,xu2022mpmr, zhou-etal-2022-melm, zhou-etal-2022-conner} under cross-lingual settings \citep{zhang-etal-2021-cross,liu-etal-2022-enhancing-multilingual, liu-etal-2022-towards-multi} can be categorized into: 
(1) feature-based methods, which generate language-independent features to facilitate cross-lingual transfer via wikification \citep{tsai-etal-2016-cross}, language alignment \citep{wu-dredze-2019-beto} or adversarial learning \citep{keung-etal-2019-adversarial}. 
(2) translation-based methods, which produce pseudo training data by translating labeled source-language data word-by-word \citep{xie-etal-2018-neural} or with the help of word alignment tools \citep{jain-etal-2019-entity, li2020unsupervised, liu-etal-2021-mulda}. 
(3) self-training methods, which generate pseudo-labeled target-language data using a model trained with labeled source-language data \citep{wu-etal-2020-single,ijcai2020-543,liang2021reinforced,chen-etal-2021-advpicker,li-etal-2022-unsupervised-multiple}.
One concurrent work \citep{ge2023prokd} that is similar to ours also aims to improve self-training for cross-lingual NER, but they adopt the traditional sequence tagging formulation, and also only apply contrastive learning on class-specific prototypes instead of actual spans.
\citet{dong2020leveraging} also leverages self-training for sentence-level cross-lingual tasks.
\paragraph{Contrastive learning} Self-supervised contrastive learning has been widely used to generate representations for various tasks \citep{pmlr-v119-chen20j, NEURIPS2020_63c3ddcc, NEURIPS2020_4c2e5eaa, NEURIPS2020_3fe23034, han-etal-2022-sancl, nguyen-etal-2022-adaptive, tan-etal-2022-domain}. 
In a nutshell, contrastive learning pulls positive pairs closer while pushing negative pairs apart. 
Supervised contrastive learning \citep{NEURIPS2020_d89a66c7} further constructs positive pairs with labeled samples of the same class, which ensures the validity of positive pairs. 
\citet{das-etal-2022-container} leverages contrastive learning for name entity recognition, but they work on monolingual few-shot settings while we focus on cross-lingual NER self-training.
\paragraph{Prototype learning} Prototype learning \citep{NIPS2017_cb8da676, DBLP:conf/iclr/WangXLF0CZ22} produces representations where examples of a certain class are close to the class-specific prototype. Several works explored prototype learning for few-shot NER \citep{10.1145/3297280.3297378, hou-etal-2020-shot, wang2022spanproto}. 

\section{Conclusions} 
In this work, we propose ContProto as a novel self-training framework for cross-lingual NER, which synergistically incorporates representation learning and pseudo label refinement.
Specifically, our contrastive self-training first generates representations where different classes are separated, while explicitly enforcing the alignment between source and target languages. 
Leveraging the class-specific representation clusters induced by contrastive learning, our prototype-based pseudo-labeling scheme further denoises pseudo labels using prototypes to infer true labels of target language spans. As a result, the model trained with more reliable pseudo labels is more accurate on the target languages.
In our method, the contrastive and prototype learning components are mutually beneficial, where the former induces clusters which makes it easier to identify the closest prototype, and the latter helps to construct more accurate sample pairs for contrastive learning.
Evaluated on multiple cross-lingual transfer pairs, our method brings in substantial improvements over various baseline methods.

\section*{Limitations}
In this work, we propose a self-training method which requires unlabeled data in target languages. Recall that we remove gold labels from readily available target-language training data from the same public NER dataset, and use them as unlabeled data in our experiments. However, this might not perfectly simulate a real-life application scenario. Firstly, most free text in target languages might not contain any predefined named entities. This requires careful data cleaning and preprocessing to produce unlabeled data ready for use. Secondly, there might be a domain shift between labeled source-language data and unlabeled target-language data, which poses a question on the effectiveness of our method.

Furthermore, the NER datasets used in this work contain only a few entity types and different entity classes are relatively balanced. However, on datasets with a larger number of classes, each class will be underrepresented in a batch and a larger batch size might be required for contrastive self-training to work satisfactorily. Also, if the entity type distribution is long-tailed, prototypes for those rare entity types might be inaccurate, and this affects the efficacy of prototype-based pseudo-labeling.

Lastly, as we observe slight drops of pseudo label quality at the end of training for some languages, the pseudo label update strategy can be refined for further improvement.

\section*{Acknowledgements}
This research is supported (, in part,) by Alibaba Group through Alibaba Innovative Research (AIR) Program and Alibaba-NTU Singapore Joint Research Institute (JRI), Nanyang Technological University, Singapore.

\pagebreak

\bibliography{anthology,custom}
\bibliographystyle{acl_natbib}











\end{document}